\documentclass[runningheads]{llncs}
\usepackage{amssymb}
\usepackage{hyperref}
\usepackage{stmaryrd}
\usepackage{graphics}
\usepackage{epsfig}
\usepackage{amsmath,amssymb}

\usepackage{multirow}
\usepackage{mathabx}

\usepackage{subfigure}
\usepackage{graphicx}
\usepackage{amsmath}
\usepackage{color}
\usepackage{arydshln}
\usepackage{rotating}
\begin{document}
\title{ ICMSC: Intra- and Cross-modality Semantic Consistency for Unsupervised Domain Adaptation on Hip Joint Bone Segmentation }
\author{
Guodong Zeng\inst{1},
Till D. Lerch\inst{2, 3},
Florian Schmaranzer\inst{2,3},
Guoyan Zheng\inst{4,*},
Juergen Burger\inst{1},
Kate Gerber\inst{1},
Moritz Tannast\inst{5},
Klaus Siebenrock\inst{2},
Nicolas Gerber\inst{1}
\\[-3.0ex]
}
\institute{ 
  sitem Center for Translational Medicine and Biomedical Entrepreneurship, University of Bern, Switzerland \ \\
  \and 
  Department of Orthopaedic Surgery, Inselspital, University of Bern, Switzerland \ \\
  \and Department of Diagnostic, Interventional and Paediatric Radiology, Inselspital, University of Bern, Switzerland \ \\
  \and Institute of Medical Robotics, School of Biomedical Engineering, Shanghai Jiao Tong Unviersity, China \ \\
  \and Department of Orthopaedic Surgery and Traumatology, University Hospital of Fribourg,  Switzerland \ \\
  \texttt{*Correspondence: guoyan.zheng@sjtu.edu.cn}
}


%
\maketitle              
\begin{abstract}

Unsupervised domain adaptation (UDA) for cross-modality medical image segmentation has shown great progress by domain-invariant feature learning or image appearance translation. Adapted feature learning usually cannot detect domain shifts at the pixel level and is not able to achieve good results in dense semantic segmentation tasks. Image appearance translation, e.g. CycleGAN, translates images into different styles with good appearance, despite its population, its semantic consistency is hardly to maintain and results in poor cross-modality segmentation. In this paper, we propose intra- and cross-modality semantic consistency (ICMSC) for UDA and our key insight is that the segmentation of synthesised images in different styles should be consistent. Specifically, our model consists of an image translation module and a domain-specific segmentation module. The image translation module is a standard CycleGAN, while the segmentation module contains two domain-specific segmentation networks. The intra-modality semantic consistency (IMSC) forces the reconstructed image after a cycle to be segmented in the same way as the original input image, while the cross-modality semantic consistency (CMSC) encourages the synthesized images after translation to be segmented exactly the same as before translation. Comprehensive experimental results on cross-modality hip joint bone segmentation show the effectiveness of our proposed method, which achieves an average DICE of 81.61\% on the acetabulum and 88.16\% on the proximal femur, outperforming other state-of-the-art methods. It is worth to note that without UDA, a model trained on CT for hip joint bone segmentation is non-transferable to MRI and has almost zero-DICE segmentation.


\end{abstract}
\section{Introduction}

Due to increasing life expectancy and an aging population, osteoarthritis is expected to be the fourth most common cause of disability and it happens very often at hip joint  \cite{murray1996global}. Femoroacetabular impingement (FAI) and developmental dysplasia of the hip (DDH) are the two main causes of hip osteoarthritis in young and active patients. The segmentation of hip joint bone structures (proximal femur and acetabulum) is an essential prior step for the diagnosis and surgical planning for FAI and DDH \cite{tannast2007noninvasive}. Manual segmentation is tedious and time-consuming, thus automatic and accurate hip joint bone segmentation is strongly demanded in clinical practice. Deep Convolutional Neural Network (DCNN) is a powerful tool for medical image analysis \cite{shen2017deep} and has been applied for hip joint bone segmentation \cite{zeng20173d}. However, it is data-hungry and often results in severe performance losses when testing on domain-shifted datasets. For example, the model trained on hip joint CT images can hardly achieve good segmentation on MR images because of the dramatic shift in appearance. A model that can be trained with annotations from only one modality but can be generalized to other modalities without the need for additional annotations is therefore highly desirable.

Unsupervised Domain Adaptation (UDA) addresses this challenge by transferring knowledge from a domain with labels (source domain) to another domain without labels (target domain), which can greatly reduce the workload for manual annotation and is especially important in the medical field which requires specialized knowledge. UDA on cross-modality medical image segmentation has shown tremendous progress by feature adaptation or image appearance adaptation. Feature adaptation methods typically learn domain invariant features by adversarial training strategy \cite{tzeng2017adversarial,dou2018unsupervised}. Specifically, a discriminator classifies which domain is the feature map from and it is used to force the segmentation network output indistinguishable features from both domains. But adapted feature learning usually cannot detect domain shifts at the pixel level and is not able to achieve good results in dense semantic segmentation tasks. Image appearance adaptation methods implement a pixel-to-pixel image translation between unpaired image domains, and the translated source-like images from the target domain can be directly tested on the models trained on the source domain \cite{chen2018semantic,chen2019unsupervised,hiasa2018cross}. Fusion of feature adaptation and image appearance adaptation is also investigated and has been applied to the cross-modality segmentation of cardiac tissue \cite{chen2020unsupervised}. Most of the image appearance adaptation models share the same spirit with CycleGAN\cite{zhu2017unpaired}, in which a forward cycle and a backward cycle are formed by translating the synthesised image back into a reconstruction image that approximates the original input image. Though they usually output source-like or target-like images in good appearance, there is no guarantee to keep semantic consistency after image translation and thus leading to poor segmentation, which is qualitatively demonstrated in Fig. \ref{fig:synthesis quality}. 

In this paper, we propose Intra- and Cross-modality Semantic Consistency (ICMSC) unsupervised domain adaptation and applied it to cross-modality hip joint bone segmentation (from CT to MRI). Our key insight is that the semantics should be consistent during image appearance translation. Specifically, our model consists of an image translation module and a domain-specific segmentation module. The image translation module consists of two generators and two discriminators like CycleGAN, while the segmentation module includes two domain-specific segmentation networks. The Intra-modality Semantic Consistency (IMSC) forces the reconstructed image after a cycle to be segmented in the same way as the original input image, while the Cross-modality Semantic Consistency (CMSC) encourages the synthesized images after translation to be segmented exactly the same as before translation. The proposed ICMSC enforces the domain-invariant semantic structure to be well preserved in the image appearance translation process, thus improving the performance of cross-modality segmentation. It should be noted that CyCADA \cite{hoffman2018cycada} also investigates semantic consistency during image translation, but significant differences exist between our work and theirs: they pre-trained a classifier on the source domain and use this classifier to enforce consistency between input and translated fake images. This only works when the pre-trained classifier can perform good segmentation on the translated fake images, but this is difficult to realise when the domain shift is dramatic like from hip joint CT to MRI. In contrast, we used domain-specific segmentation networks, one for the source domain and one for the target domain, which is more promising to have good segmentation for both input and translated images in different domains. Further, we have IMSC while they don't. In addition, our method integrates the unsupervised image translation and domain-specific segmentation in one unified model, which can be trained in an end-to-end manner. Experimental results also demonstrate that our method outperformed theirs. 

The main contributions of this paper are summarized as follows: (1) We present an elegant and unified domain adaptation framework for cross-modality segmentation of hip joint from CT to MRI, which can significantly reduce the workload for data annotation. (2) We propose intra- and cross-modality semantic consistency (ICMSC) for unsupervised domain adaptation as additional supervisory objectives to enforce semantic consistency during image translation, thus improving cross-modality segmentation performance. (3) We validate the effectiveness of our method in the challenging task of cross-modality hip joint segmentation from CT to MRI, and our approach outperforms the state-of-the-art methods.

\begin{figure}[!ht]
\centering
\includegraphics[width=\textwidth]{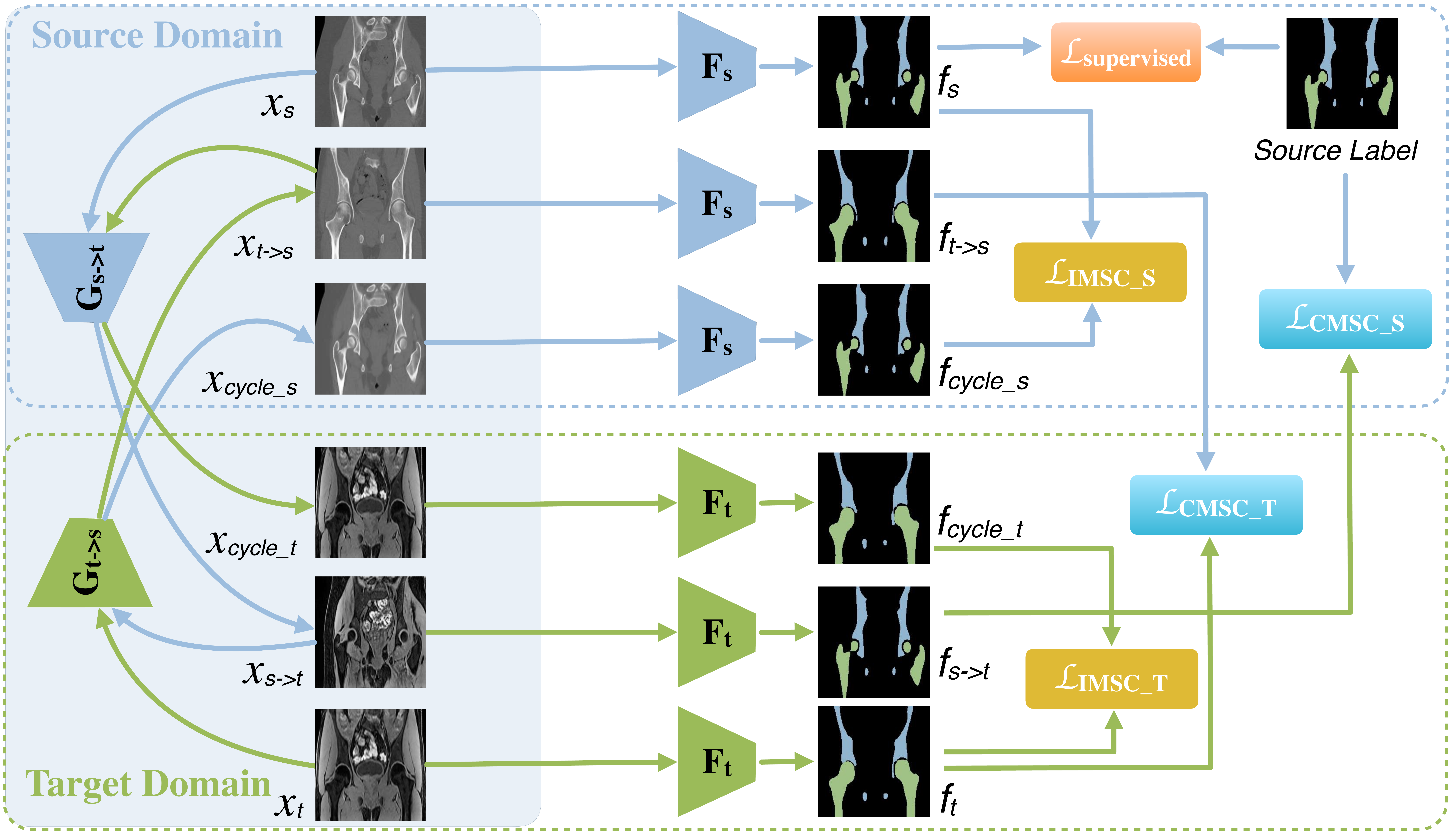}
\caption{ Method overview. Our method consists of an image translation module (left side in shadow) and a domain-specific segmentation module (right side). The image appearance translation module is a typical CycleGAN  which contains two generators  ($G_{s->t}$, $G_{t->s}$) and two discriminators ($D_s$, $D_t$). The two discriminators are omitted in this figure for clarity. The blue arrows and green arrows represent the source and target domain, respectively. The segmentation module is composed of one segmentation network for the source domain ($F_s$) and another segmentation network for the target domain ($F_t$). All real input, translated and reconstructed cycle images from image translation module are fed into corresponding domain-specific segmentation network for prediction.  Our contribution is the proposed Intra-modality Semantic Consistency loss ($\mathcal{L}_{IMSC\_S}$ in the source domain and $\mathcal{L}_{IMSC\_T}$ in the target domain) and Cross-modality Semantic Consistency loss ($\mathcal{L}_{CMSC\_S}$ in the source domain and $\mathcal{L}_{CMSC\_T}$ in the target domain) to encourage semantic consistency during image translation. }

\label{fig:overview}
\end{figure}

\section{Method}


Our method addresses the challenge of cross-modality UDA for hip joint bone segmentation from CT to MRI. Formally, we have dataset $\{X_s, Y_s, X_t\}$, where $X_s$ is the source image set, $Y_s$ is the source label set, and $X_t$ is the unlabelled target image set, but we have no access to the target label set $Y_t$. Let $x_s$, $y_s$ , $x_t$ and $y_t$ represent a single image or label from $X_s$, $Y_s$, $X_t$ and $Y_t$, respectively. Our goal is to learn a model which can perform accurate segmentation on the target images $x_t$ without using the the target labels $y_t$. 

\subsection{Overview of Our Method. }  

As shown in Fig. \ref{fig:overview},  our framework consists of an image appearance translation module (left side in shadow) and a domain-specific segmentation module (right side). The image appearance translation module translates images between domains by pixel-to-pixel mapping in an un-supervised way. It adopts an architecture like CycleGAN \cite{zhu2017unpaired}, which contains two generators ($G_{s->t}$, $G_{t->s}$) and two discriminators ($D_s$, $D_t$). In the forward cycle (in blue arrows), the source input image $x_s$ is translated into target-like image $x_{s->t} = G_{s->t}(x_s)$, and reconstructed as a cycle source image $x_{cycle\_s}=G_{t->s}(x_{s->t})$. Similarly, the target input image $x_t$ is translated into source-like image $x_{t->s} = G_{t->s}(x_t)$ and reconstructed as a cycle target image $x_{cycle\_t}=G_{s->t}(x_{t->s})$ in the backward cycle ( in green arrows). The domain-specific segmentation module includes two segmentation networks, i.e. one for the source domain ($F_s$) and another for the target domain ($F_t$). The input, translated and reconstructed images from both domains are fed into the corresponding domain segmentation networks in the domain-specific segmentation module for prediction. 

To encourage the domain-invariant semantic structure to be consistent in the image translation process, we propose the IMSC loss $\mathcal{L}_{IMSC}$ and CMSC loss ($\mathcal{L}_{CMSC}$). The main motivation is that synthesised image after translation and reconstruction should be segmented in the same way as the original input image.  Specifically, $\mathcal{L}_{IMSC}$ forces the same segmentation between the original input and reconstructed cycle image in the same modality, which can be represented as $F_s(x_s) \approx F_s(x_{cycle\_s})$ and $F_t(x_t) \approx F_t(x_{cycle\_t})$. On the other hand, $\mathcal{L}_{CMSC}$ constrains the same segmentation between the images before and after translation in different modalities, which is formally displayed as $F_s(x_s) \approx F_t(x_{s->t})$ and $F_t(x_t) \approx F_s(x_{t->s})$. Since the source label $y_s$ already exists, we replace  $F_s(x_s)$ directly with the source labels $y_s$ in the cross-modality semantic consistency. In the testing phase, target MR images are directly fed into $F_t$ to get segmentation.

\subsection{Learning Objective of Our Method. } 

The overall training objective of our method includes five loss terms. Specifically, the first loss term $\mathcal{L}_{supervised}$ is the supervised segmentation loss on the source domain, which guides the source domain segmentation network $F_s$ to perform accurate segmentation on given source images $x_s$. The second image domain adversarial loss $\mathcal{L}_{adv}^{img}$ and the third image reconstruction loss $\mathcal{L}_{rec}^{img}$ are used for the training of un-supervised image appearance translation. The image domain adversarial loss distinguishes whether the images look like real or not, and this can force the generators to output images which are indistinguishable with the real images. The reconstruction loss $\mathcal{L}_{rec}^{img}$ enforces the generators to perform self-reconstruction between the input image and reconstructed cycle  image. The fourth loss is the intra-modality semantic consistency loss $\mathcal{L}_{ICSC}$, which forces that the segmentation between input and reconstructed cycle image to be the same. The last loss is the cross-modality semantic consistency loss $\mathcal{L}_{CMSC}$, which constrains the input image and translated images across domain to have the same segmentation from corresponding domain segmentation network. The overall training loss function is defined as:

\begin{equation}
    \begin{split}
         \mathcal{L}_{ICMSC} = \mathcal{L}_{supervised} + \lambda_{adv} \cdot \mathcal{L}_{adv}^{img} + \lambda_{rec} \cdot \mathcal{L}_{rec}^{img}   \quad \\
           +  \lambda_{intra} \cdot \mathcal{L}_{IMSC} + \lambda_{cross} \cdot \mathcal{L}_{CMSC}
    \end{split}
    \label{equation:overall loss}
\end{equation}

\subsubsection{Supervised Segmentation Loss on Source Domain.}
We first define a general supervised loss between the predicted probability map and ground truth segmentation, no matter which domain is the image $x$ from.  A segmentation network $F$ takes an image $x\subset R^{H\times W }$ as input and outputs probability map after the last Softmax layer: $F(x) = P_x \subset R^{H\times W \times C}$, where $H,W,C$ represent the height, the width of image and the class number of segmentation, respectively. The probability of each pixel for c-th class is represented as $p^{(h,w,c)}_{F(x)}$. The supervised loss function mixes the cross entropy loss and the dice coefficient loss \cite{milletari2016v}, and is defined as:

\begin{equation}
    \begin{split}
        {\mathcal{L}_{seg(x, y, F)}} = -\sum_{h=1}^H \sum_{w=1}^W \sum_{c=1}^C  y^{(h,w,c)} \cdot  log(p^{(h,w,c)}_{F(x)}) \quad\quad\quad\quad\quad\quad \\
           -\lambda_1 \sum_{h=1}^H \sum_{w=1}^W \sum_{c=1}^C \frac{2 \cdot y^{(h,w,c)} \cdot \hat y^{(h,w,c)}_{F(x)}} { y^{(h,w,c)} \cdot y^{(h,w,c)} + \hat y^{(h,w,c)}_{F(x)} \cdot \hat y^{(h,w,c)}_{F(x)}}
    \end{split}
    \label{equation:supervised cross entropy and dice loss}
\end{equation}
where $y^{(h,w,c)}$ and $\hat y^{(h,w,c)}_{F(x)}$ represent the ground truth and the predicted segmentation in the form of one-hot vector with C-class, respectively. Based on the definition of general segmentation loss in Eq. \ref{equation:supervised cross entropy and dice loss}, the supervised segmentation on the source domain $\mathcal{L}_{supervised}$ is defined as:
\begin{equation}
  {\mathcal{L}_{supervised} =  \frac{1}{|X_s|} \sum_{x_s \in X_s} \mathcal{L}_{seg}(x_s,y_s,F_s)  }
  \label{equation:supervised loss on source domain }
\end{equation}

\subsubsection{Image Domain Adversarial Loss.}
The image domain adversarial loss $\mathcal{L}_{adv}^{img}$ is a combination of two adversarial losses from source and target domains, i.e. $\mathcal{L}_{adv}^{img} = \mathcal{L}_{D_s} + \mathcal{L}_{D_t} $. 
Two discriminators, $D_s$ and $D_t$, aim to distinguish the real input and fake translated images in their corresponding domain. Specifically, $D_s$ takes the real source image $x_s$ and translated source-like image $x_{t->s}$ as input and classifies them (label 1 for source $x_s$ and 0 for $x_{t->s}$). Let $\mathcal{L}_D$ be the cross-entropy loss. We can then define $\mathcal{L}_{D_s}$ and $\mathcal{L}_{D_t}$ as:
\begin{equation}
    \begin{split}
        \mathcal{L}_{D_s} = \frac{1}{|x_s|} \sum_{x_s \in X_s} \mathcal{L}_D(D_s(x_s),1) +   \frac{1}{|x_t|} \sum_{x_t \in X_t} \mathcal{L}_D(D_s(G_{t->s}(x_t)),0)   \\
        \mathcal{L}_{D_t} = \frac{1}{|x_t|} \sum_{x_t \in X_t} \mathcal{L}_D(D_t(x_t),1) +   \frac{1}{|x_s|} \sum_{x_s \in X_s} \mathcal{L}_D(D_t(G_{s->t}(x_s)),0)  
    \end{split}
    \label{equation:image domain adversarial loss}
\end{equation}

\subsubsection{Image Reconstruction Loss.}
The image reconstruction loss $\mathcal{L}_{rec}^{img}$ between the input and reconstructed cycle image is used to encourage the preserve of image content during the translation process. The motivation lies in that we should achieve the same image after a cycle: $G_{t->s}(G_{s->t}(x_s)) \approx x_s$ for any source image $x_s$ and $G_{s->t}(G_{t->s}(x_t)) \approx x_t$ for any target image $x_t$. Formally, the image reconstruction loss is defined as:

\begin{equation}
    \begin{split}
        \mathcal{L}_{rec}^{img} = \frac{1}{|x_s|} \sum_{x_s \in X_s} [|| G_{t->s}(G_{s->t}(x_s)) - x_s ||_1] \\
        	+   \frac{1}{|x_t|} \sum_{x_t \in X_t} [|| G_{s->t}(G_{t->s}(x_t)) - x_t ||_1]
    \end{split}
    \label{equation:reconstruction cross entropy and dice loss}
\end{equation}

\subsubsection{Intra-modality Semantic Consistency (IMSC) Loss.}

Though L1-based image reconstruction loss $\mathcal{L}_{rec}^{img}$ has been applied to preserve image content, there is no guarantee for the consistent semantics after a cycle reconstruction. This motivates us to propose the IMSC loss $\mathcal{L}_{ICSC}$, which enforces the segmentation between input and reconstructed cycle image should be exactly the same from both domains, i.e. IMSC loss in source domain $\mathcal{L}_{ICSC\_S}$, and IMSC loss in target domain $\mathcal{L}_{ICSC\_T}$.  In other words, we want $F_s(x_s) \approx F_s(G_{t->s}(G_{s->t}(x_s)))$ for any $x_s$ and $F_t(x_t) \approx F_t(G_{s->t}(G_{t->s}(x_t)))$ for any $x_t$, which is defined as:

\begin{equation}
    \begin{split}
        \mathcal{L}_{IMSC} &= \mathcal{L}_{IMSC\_S} + \mathcal{L}_{IMSC\_T} \quad\quad\quad\quad\quad\quad\quad\quad\quad\quad\quad\quad \\
        &= \frac{1}{|x_s|} \sum_{x_s \in X_s} \mathcal{L}_{seg}(G_{t->s}(G_{s->t}(x_s)), F_s(x_s), F_s) \\
        \quad  &+ \frac{1}{|x_t|} \sum_{x_t \in X_t} \mathcal{L}_{seg}(G_{s->t}(G_{t->s}(x_t)), F_t(x_t), F_t) 
    \end{split}
    \label{equation:intra modality  loss}
\end{equation}

\subsubsection{Cross-modality Semantic Consistency (CMSC) Loss.}

Though $\mathcal{L}_{ICSC}$ has been used to enforce the semantic consistency between the input and reconstructed cycle image, the semantic consistency between the input and translated image across domain is still not guaranteed. This motivates us to propose the CMSC loss $\mathcal{L}_{CMSC}$, which constrains that the segmentation after translation will be exactly the same as before translation from both domains, i.e. CMSC loss in source domain $\mathcal{L}_{CMSC\_S}$, and CMSC loss in target domain $\mathcal{L}_{CMSC\_T}$. In other words, we want $F_s(x_s) \approx F_t(G_{s->t}(x_s))$ for any source image $x_s$ and $F_t(x_t) \approx F_s(G_{t->s}(x_t))$, which is defined as:

\begin{equation}
    \begin{split}
        \mathcal{L}_{CMSC} &=  \mathcal{L}_{CMSC\_S} + \mathcal{L}_{CMSC\_T}\\
        	&= \frac{1}{|x_s|} \sum_{x_s \in X_s} \mathcal{L}_{seg}(G_{s->t}(x_s), y_s, F_t) \\
        	&+  \frac{1}{|x_t|} \sum_{x_t \in X_t} \mathcal{L}_{seg}(x_t, F_s (G_{t->s}(x_t)), F_t) 
    \end{split}
    \label{equation:cross modality loss}
\end{equation}

Based on above detaily described each loss term in $\mathcal{L}_{ICMSC}$, our aim is to train a segmentation network on the target domain ($F_t$) by optimizing the min-max game as below: 

\begin{equation}
    F_t^* = arg \operatorname*{min}_{F_t} \operatorname*{min}_{ \substack{F_s \\  G_{s->t} \\ G_{t->s}} } \operatorname*{max}_{D_s, D_t} \mathcal{L}_{ICMSC}
    \label{equation:final optimisation}
\end{equation}

\subsection{Implementation Details.}
\label{Sec:networkarchitecture}
The proposed method was implemented in Pytorch and trained with a GTX 1080 Ti graphics card. For the image appearance translation module, we use the same architecture as standard CycleGAN \cite{zhu2017unpaired}. The two segmentation networks in the domain-specific segmentation module ($F_s, F_t$) adopt the same architecture in \cite{zeng2020entropy}, which consists of 3 convolution layers, 4 dilated residual blocks, again 3 convolution layers, a deconvolution layer and a final Softmax layer. Each convolution layer is immediately followed by a batch normalization (BN) and a ReLU layer. We trained our network in total $120$ epochs from scratch, and the batch size was $1$. In the first 60 epochs, the network was only trained with the image domain adversarial loss $\mathcal{L}_{adv}^{img}$ and image reconstruction loss $\mathcal{L}_{rec}^{img}$ for unsupervised image translation. In the second 60 epochs, the supervised segmentation loss in the source domain and our proposed $\mathcal{L}_{IMSC}$ and $\mathcal{L}_{CMSC}$ were included for training. We used the Adam optimiser with an initial learning rate of $2\times10^{-4}$ for the image appearance translation module. For segmentation networks, the initial learning rate was $1\times10^{-3}$ and then linearly reduced to 0. We have $\lambda_{adv} = 1.0 $, $\lambda_{rec} = 10.0$, $\lambda_{intra} = 0.1$,  $\lambda_{cross} = 0.1$ so that each weighed loss term will be in similar range.

\section{Experiments and Results}

\subsection{Dataset and Preprocessing.}

The effectiveness of the proposed method was evaluated using cross-modality hip joint bone segmentation. The dataset we used consists of unpaired 29 CT and 19 MRI volumes collected by University Hospital X (anonymized for review).  Experienced clinicians manually segmented the hip joint bones from both CT and MRI as ground truth. We used the CT as source domain and MRI as target domain. Our goal is to train a segmentation model that is trained with labeled CT dataset and unlabelled MRI, but can perform accurate segmentation on the hip joint MRI. Note that ground truth segmentation from MRI was not used during training. For all experiments described below we used the Dice Coefficients (DICE) and the Average Surface Distance (ASD) as evaluation metrics.

For both source and target dataset, one third of the total number of images was held-out for testing, the others were used for training and validation. Specifically, the 29 CT images were split into 15 for training, 3 for validation and 9 for testing. We also split the 19 MRI images from the target domain into 10 for training, 2 for validation and 7 for testing. All CT and MRI images were resampled to a voxel spacing of $1.25 \times 1.25 \times 1 \  mm^3  $ and then cropped to $300 \times 160 \times 240 $ voxels around the midpoint of two proximal femoral head centers, which helps to reduce unnecessary calculation on unrelated background.  In total, we have $15 \times 160 = $2400 slices from CT and $10 \times 160 = $ 1600 slices from MRI for training. For the test on the target domain, $7 \times 160 = $ 1120 slice images were used. Each image is linealy rescaled to [-1,1] before fed into the neural network. Augmentation strategies such as random flipping, contrast adjustment, and scaling were used to enlarge the training data set and reduce over-fitting.

\subsection{Experimental Results.}

\subsubsection{Comparison with the State-of-the-art Methods. } We compared our method with state-of-the-art methods: ADDA \cite{tzeng2017adversarial}, CycleGAN \cite{zhu2017unpaired}, CyCADA \cite{hoffman2018cycada} and SIFA \cite{chen2020unsupervised}. Among these four methods, ADDA use feature adaptation only, CycleGAN adapts image appearance only, while CyCADA and SIFA employ both feature and image appearance adaptation. We directly use the code from the paper if they are available, otherwise we re-implement it for our cross-modality hip joint bone segmentation task. For CycleGAN, the code from the original paper was used for unsupervised image translation. Afterwards, the target MRI were then translated into CT-like images and directly tested on pre-trained CT segmentation model. We also compared our results with the lower bound, where no adaptation from the pre-trained CT segmentation model was performed, and with the upper bound, where the model was trained directly with the labelled target dataset.

\begin{table}[tbp!]
\centering
\caption{ Quantitative comparative results between our method and other state-of-the-art UDA methods for the task of cross-modality hip joint bone segmentation. The best results are highlighted with bold font.}
\label{Tab:resultsWithSOTA}
\resizebox{\textwidth}{!}
{

\begin{tabular}{l|l|c|c|c|c}
\hline
\multicolumn{1}{c|}{\multirow{2}{*}{\textbf{Methods}}} & \multicolumn{1}{c|}{\multirow{2}{*}{\textbf{Adaptation}}} & \multicolumn{2}{c|}{\textbf{DICE (\%)  $\uparrow$}}            & \multicolumn{2}{c}{\textbf{ASD (mm) $\downarrow$}}        \\ \cline{3-6} 
\multicolumn{1}{c|}{}                                  & \multicolumn{1}{c|}{}                                     & \textbf{Acetabulum} & \textbf{ Femur} & \textbf{Acetabulum} & \textbf{ Femur} \\ \hline
No Adaption                                            & Train wtih Source Label Only                              & 0.08                & 0.01                    & 29.36               & 31.27                   \\ \hline
ADDA \cite{tzeng2017adversarial}                                                  & Feature Align                                        & 50.56               & 64.80                   & 4.20                & 4.87                    \\
CycleGAN  \cite{zhu2017unpaired}                                                 & Image Appearance Adapt                               & 74.55               & 72.78                   & 2.05                & 9.16                    \\
CyCADA \cite{hoffman2018cycada}                                                 & Appearance Adapt + Feature Align                & 73.80               & 83.18                   & 3.18                & 2.24                    \\
SIFA  \cite{chen2020unsupervised}                                                 & Appearance Adapt + Feature Align                & 79.60               & 83.32                   & 2.93                & 2.67                    \\
Ours                                                   & Appearance Adapt + ICMSC              & \textbf{81.61}      & \textbf{88.16}          & \textbf{1.61}       & \textbf{1.42}           \\ \hline
Target Model                                           & Train with Target Label Directly                          & 95.23               & 97.29                   & 0.43                & 0.35                    \\ \hline
\end{tabular}
}
\end{table}

\begin{figure}[tbp!]
\centering
\includegraphics[width=\textwidth]{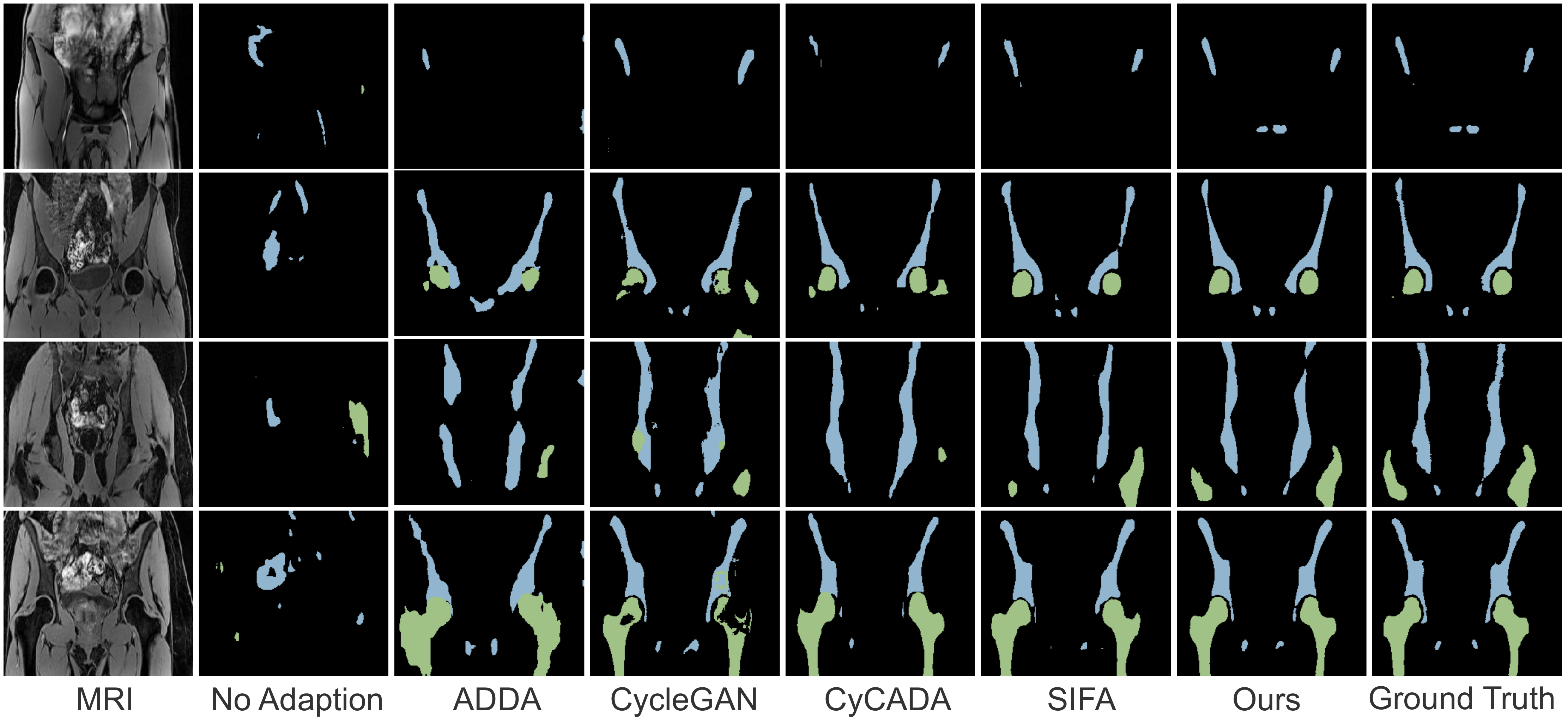}
\vspace*{-5mm}
\caption{Qualitative comparison of segmentation results by different methods. The blue and green colors are corresponding to the acetabulum and proximal femur segmentation, respectively. Each row is one example. } 
\label{fig:qualitativeComparison}
\end{figure}

The quantitative comparison results are displayed in Tab. \ref{Tab:resultsWithSOTA}. The first row shows the results from no adaptation, which achieved an average DICE of 0.08\% and 0.01\% for the acetabulum and proximal femur, respectively. This illustrates the dramatic domain shift between hip joint CT and MRI images. Remarkably, our method yielded an average DICE of 81.61\% and 88.16\%, an average ASD of 1.61 mm and 1.41 mm for the acetabulum and proximal femur, respectively. Our method outperformed ADDA, CycleGAN, CyCADA and SIFA in all metrics, indicating the advantage of our proposed ICMSC in maintaining semantic consistency.  A visual comparison of the segmentation results from different methods is shown in Fig. \ref{fig:qualitativeComparison}. We can observe that the results from no adaptation in the second column are very noisy and cluttered. They cannot predict any correct segmentation, due to the dramatic domain shift between hip joint CT and MRI,  By using feature alignment or image adaptation alone, ADDA and CycleGAN output better results in the third and fourth column, but the segmentation is still very poor. CyCADA and SIFA take the benefits of combining feature alignment and image appearance adaptation, which provide meaningful predictions about the overall structure, but cannot not maintain the semantic consistency in the local structure or morphology that corresponds to the ground truth segmentation. For example, in the last row, the segmentation of the left femur in CyADDA and the segmentation of the right femur in SIFA show obvious spurs at greater trochanteric regions while the input MRI and ground truth segmentation do not. In comparison, our method is able to provide predictions that are closer to the ground truth in both global and local structures.

\subsubsection{ Ablation Study: Effectiveness of Key Components. } 

To investigate the role of different semantic consistency losses, we conducted a study to compare five variants of our method. The experimental results are presented in Tab. \ref{Tab:ablation study results}. The baseline is a typical CycleGAN and then four semantic consistency losses were added subsequently. When the first loss $\mathcal{L}_{IMSC\_S}$ is added, the increase in DICE from 72.78\% to 79.32\% on the proximal femur over the baseline, clearly demonstrates the importance of IMSC after a cycle in the source domain.Further, the inclusion of $\mathcal{L}_{CMSC\_S} $, which achieves an average DICE of 78.46\% for the acetabulum and 82.57\% for the proximal femur, is achieved by enforcing the CMSC after the source images are translated into target-like images. Then $\mathcal{L}_{IMSC\_T}$ is added to encourage IMSC after a cycle on the target domain. Finally, $\mathcal{L}_{CMSC\_S}$ encourages the CMSC after the target images are translated into source-like images, leading to futher obvious improvement.

\subsubsection{ Ablation Study: Image Translation Quality without ICMSC. } 
\begin{figure}[tbp!]
    \centering
    \includegraphics[width=\textwidth]{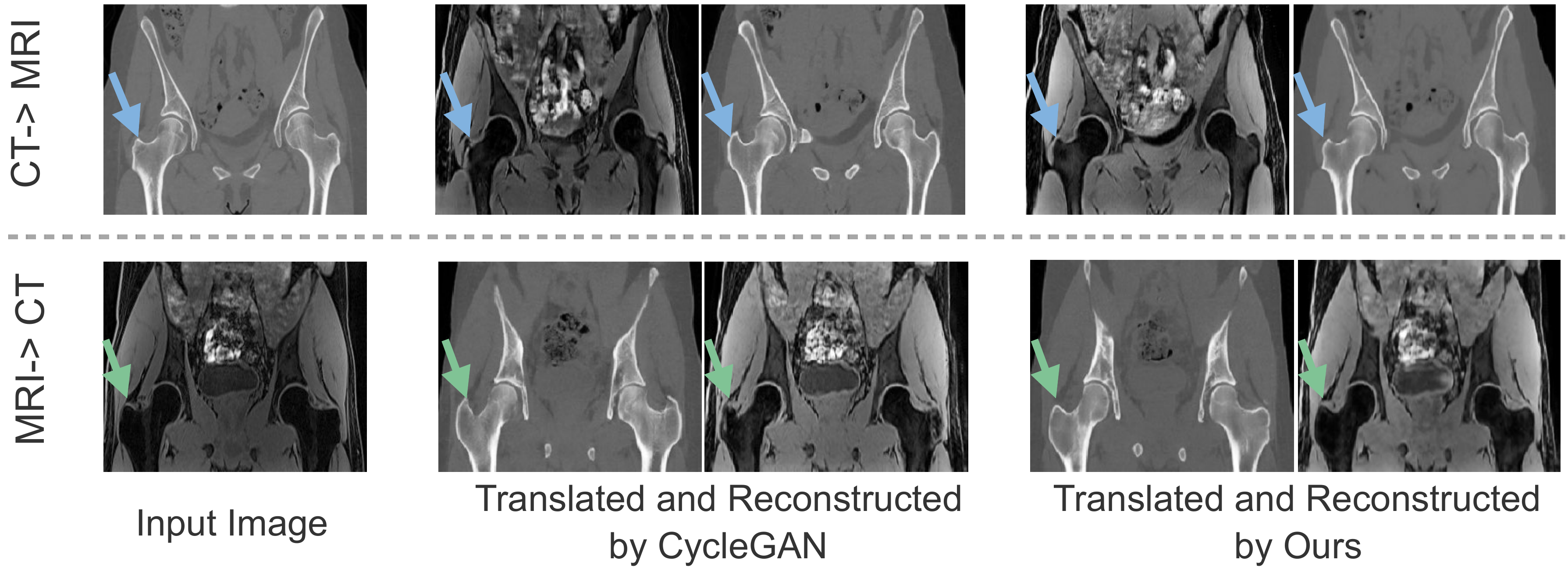}
    \setlength{\belowcaptionskip}{-10pt}
    \vspace*{-8mm}
    \caption{Qualitative comparison of unsupervised image translation between CycleGAN and ours. The shape of the greater trochanter in the proximal femur, to which the blue and green arrows point, is not consistent during the image translation and reconstruction process in CycleGAN, while our method maintains the semantic consistency.  }
    \label{fig:synthesis quality}
\end{figure}

We compare the unsupervised image translation with and without our proposed ICMSC, and the visual comparison is shown in Fig. \ref{fig:synthesis quality}. The first row shows image translation from CT to MRI and reconstruction back to CT, while the second line shows image translation from MRI to CT and reconstruction back to MRI. We note that although the standard CycleGAN outputs source and target-like images in good appearance, semantic consistency is not guaranteed. For example, the shape of the greater trochanter in the proximal femur, which is located where the blue and green arrows point, has been altered and cannot maintain consistency. The translated CT-like images show the mismatched semantic structure as the original MRI, which will lead to incorrect segmentation prediction. The method we propose uses two domain-specific segmentation networks to enforce semantic consistency during the image translation and reconstruction process. Qualitatively, we find that femoral shapes are consistent before and after translation and reconstruction.

\begin{table}[tbp!]
\centering
\caption{ Quantitative comparison of five different variants of our method for cross-modality hip joint bone segmentation. The best results are highlighted with bold font.}
\label{Tab:ablation study results}
\resizebox{\textwidth}{!}
{
    \begin{tabular}{l|ccccc|c|c}
    \hline
    \multicolumn{1}{c|}{\multirow{2}{*}{\textbf{Methods}}} & \multirow{2}{*}{\textbf{$\mathcal{L}_{rec}^{img} +  \mathcal{L}_{adv}^{img}$  } } & \multirow{2}{*}{\textbf{ $\mathcal{L}_{IMSC\_S} $ }} & \multirow{2}{*}{\textbf{ $\mathcal{L}_{CMSC\_S} $  }} & \multirow{2}{*}{\textbf{ $\mathcal{L}_{IMSC\_T} $  }} & \multirow{2}{*}{\textbf{ $\mathcal{L}_{CMSC\_T} $  }} & \multicolumn{2}{c}{\textbf{DICE (\%)  $\uparrow$}}   \\ \cline{7-8} 
    \multicolumn{1}{c|}{}                                  &                                          &                                          &                                             &                                          &                                          & \textbf{ Acetabulum } & \textbf{ Femur } \\ \hline
    No Consistency                                         & $\checkmark$                                &                                          &                                             &                                          &                                          & 74.55               & 72.78          \\
    + Intra Consistency (Source)              & $\checkmark$                                & $\checkmark$                                &                                             &                                          &                                          &     74.97                &       79.32         \\
    + Cross Consistency (Source)                     & $\checkmark$                                & $\checkmark$                                & $\checkmark$                                   &                                          &                                          &  78.46                   &    82.57            \\
    + Intra Consistency (Target)                 & $\checkmark$                                & $\checkmark$                                & $\checkmark$                                   & $\checkmark$                                &                                          & 76.93               & 84.94          \\
    + Cross Consistency (Target)                    & $\checkmark$                                & $\checkmark$                                & $\checkmark$                                   & $\checkmark$                                & $\checkmark$                                & \textbf{81.61}               &  \textbf{88.16}          \\ \hline
    \end{tabular}
}
\vspace*{-3mm}
\end{table}

\section{Conclusion and Discussion\textbf{}}
In summary, we presented an UDA framework that unifies image appearance translation and domain-specific segmentation. By incorporating intra- and cross-modality semantic consistency (ICMSC), our method can maintain better semantic consistency for image translation, and thus improve the performance of cross-modality segmentation. Comprehensive experimental results on cross-modality segmentation of the hip joint from CT to MRI demonstrate the effectiveness of our proposed method. Notably, our method can be elegantly trained end-to-end and easily extended to other UDA tasks. A limitation is that our method is memory intensive, since it involves several sub-networks during training. Therefore, we use a 2D neural network for the 3D MRI segmentation. Our future work will investigate how to reduce the required GPU memory and how to integrate 3D information.


%
%
\label{sect:bib}
\bibliographystyle{splncs}
\bibliography{ref}
\end{document}